\documentclass{article}

\usepackage[most]{tcolorbox}
\usepackage{PRIMEarxiv}
\usepackage{amsmath}
\usepackage[utf8]{inputenc} 
\usepackage[T1]{fontenc}    
\usepackage{hyperref}       
\usepackage{url}            
\usepackage{booktabs}       
\usepackage{amsfonts}       
\usepackage{nicefrac}       
\usepackage{microtype}      
\usepackage{lipsum}
\usepackage{fancyhdr}       
\usepackage{graphicx}       
\graphicspath{{media/}}     
\usepackage{lineno} 
\usepackage{wrapfig}
\title{Diagnostics for Individual-Level Prediction Instability in Machine Learning for Healthcare}

\author{
  Elizabeth W. Miller \\
  School of Data Science \\
  University of Virginia \\
  Charlottesville\\
  \texttt{\/zrc3hc@virginia.edu} \\
   \And
  Jeffrey D. Blume \\
  School of Data Science \\
  University of Virginia \\
  Charlottesville\\
  \texttt{blume@virginia.edu} \\
}

\usepackage{natbib}
\begin{document}
\maketitle

\begin{abstract}
 In healthcare, predictive models increasingly inform patient-level decisions, yet little attention is paid to the variability in individual risk estimates and its impact on treatment decisions. For overparameterized models, now standard in machine learning, a substantial source of variability often goes undetected. Even when the data and model architecture are held fixed, randomness introduced by optimization and initialization can lead to materially different risk estimates for the same patient. This problem is largely obscured by standard evaluation practices, which rely on aggregate performance metrics (e.g., log-loss, accuracy) that are agnostic to individual-level instability. As a result, models with indistinguishable aggregate performance can nonetheless exhibit substantial procedural arbitrariness, undermining clinical trust and decision consistency. To diagnose the problem, we propose an evaluation framework that quantifies individual-level prediction instability across repeated instantiations of an otherwise fixed learning pipeline. The framework introduces two complementary diagnostics: empirical prediction interval width (ePIW), which captures variability in continuous risk estimates, and empirical decision flip rate (eDFR), which measures instability in threshold-based clinical decisions. We apply these diagnostics to simulated data and 30-day mortality following myocardial infarction (GUSTO-I) dataset \citep{gusto1993thrombolytic}. Across settings, we find that for overparameterized machine-learning models, randomness arising solely from optimization and initialization can induce individual-level variability comparable to that produced by resampling the entire training dataset. Although logistic regression and neural networks often achieve similar aggregate performance, neural networks exhibit substantially greater instability in individual risk predictions, particularly near clinically relevant decision thresholds where small perturbations can alter treatment recommendations. These findings indicate that standard evaluation practices are insufficient for assessing clinical reliability and that stability diagnostics should be incorporated into routine model validation; when predictive accuracy is comparable, individual-level reliability should be a primary criterion for model selection. 
\end{abstract}

\keywords{individual-level prediction stability \and prediction uncertainty \and algorithmic randomness}

\section{Introduction}
Artificial Intelligence (AI) holds promise for transforming the healthcare system and is frequently positioned as the next frontier for clinical support \citep{poon2025adoption, atkinson2023mlhealthcare}. Yet the adoption of clinical risk stratification models remains stubbornly low: many healthcare organizations have deployed AI for tasks such as early sepsis detection, but only 38\% report high success in this area \citep{poon2025adoption}. This gap between controlled performance and real-world reliability has several well-documented explanations. Models can fail due to underspecification of the learning pipeline \citep{damour2022underspecification}, distributional shifts between training and deployment populations \citep{damour2022underspecification, pmlr-v238-bansak24a}, poor probability calibration \citep{pmlr-v70-guo17a}, biased or mislabeled training data \citep{doi:10.1126/science.aax2342, 9729424}, and sensitivity to arbitrary choices in modeling architecture or variable selection \citep{yusufujiang2026prediction, riley2023multiverse, riley2023stability}. Taken together, these challenges reflect a general unreliability in complex predictive systems when applied to high-stakes decisions.

A less visible but equally consequential source of unreliability concerns the stochastic nature of the learning pipeline itself. In classical statistical frameworks, uncertainty in model estimates is attributed primarily to sampling variability in the observed data, with the estimation procedure treated as fixed \citep{hastie2009elements}. This assumption does not apply to modern high-capacity models. Neural networks trained with stochastic optimization can converge to different solutions depending on random seed initialization alone \citep{ganesh2024differenthorses}, even when the training data, architecture, and hyperparameters are held constant. In the overparameterized regime  (i.e., where the number of model parameters far exceeds the training sample size) there exist many equally performing solutions scattered across a complex loss landscape, and the specific solution reached is as much a product of optimization randomness as of the data. This optimization-induced variability is a distinct and largely overlooked source of predictive uncertainty.

The consequences of this variability are most acutely felt at the level of individual patients, yet standard evaluation practices are not designed to detect them. Aggregate metrics such as binary cross-entropy (BCE), AUC, or accuracy characterize expected performance across a population, but average away the instability that matters for individual clinical decisions \citep{pmlr-v193-lopez-martinez22a}. A model can achieve high aggregate performance while producing a meaningfully different risk estimate, and a different treatment recommendation, for the same patient depending on which random seed was used during training. Two models that appear equivalent under standard validation may thus differ substantially in their reliability for any given individual.

To quantify these hidden sources of variability, we propose two complementary diagnostics. The first, empirical prediction interval width (ePIW), measures the dispersion in a patient's predicted risk across repeated retrainings of the model to measure how wide a continuous risk estimate can vary as a function of the learning pipeline. The second, empirical decision flip rate (eDFR), measures how frequently a patient's binary clinical recommendation changes across retrainings at a given risk threshold. We evaluate models spanning a continuum of expressive capacity, from logistic regression to its high-capacity generalization, the neural network, as a function of both variation in the training data and stochasticity in the optimization routine, focusing on settings where all models achieve comparable aggregate performance. We observe the following:
\begin{enumerate}
    \item Models with comparable aggregate performance can exhibit substantially different levels of individual-level prediction stability, with high-capacity models often producing wider dispersion in predicted risks across retrainings.
    \item Stochasticity introduced through the optimization process can induce instability in the individual-level predictions of high-capacity models that is comparable in magnitude to that arising from resampling the training data.
    \item Instability is not uniform across the risk spectrum, and variability in predicted risk does not necessarily translate into instability in downstream binary decisions.
\end{enumerate}

We apply this evaluation framework to controlled simulations and the GUSTO-I clinical dataset \citep{gusto1993thrombolytic}. Our experiments reveal that models with stronger structural assumptions, such as logistic regression, consistently exhibit greater stability than flexible neural networks, which remain highly sensitive to algorithmic randomness. These results underscore a critical trade-off between model flexibility and the reliability of individual-level predictions, raising important considerations for the deployment of high-capacity models in high-stakes clinical decision-making.

\section{Background and Related Work}
Risk stratification models are among the most consequential applications of quantitative methods in healthcare. In clinical practice, a physician may use such a model to estimate a patient's probability of a serious outcome (e.g., acute myocardial infarction) and decide based on that probability whether to initiate treatment, escalate monitoring, or discharge \citep{doi:10.1161/CIRCULATIONAHA.114.014508}. This places an unusually demanding set of requirements on clinical prediction models: they must not only discriminate well between patients at different levels of risk, but produce well-calibrated probability estimates (i.e., predicted risks that correspond to observed event rates) and consistent individual-level predictions (i.e.,  predictions for the same patient that remain relatively stable across retrainings of the model). Standard aggregate evaluation practices (e.g., AUC, log-loss) are designed to assess the first two of these requirements, but are not equipped to evaluate the third.

For a given prediction task, there can often exist a vast set of models that achieve nearly indistinguishable out-of-sample performance while relying on fundamentally different decision logic \citep{breiman2001twocultures, semenova2019simpler}. The transition toward overparameterized models, where the parameter count ($p$) exceeds the training sample size ($n$), has exponentially expanded this space of competitive but divergent solutions \citep{belkin2019biasvariance, hastie2022ridgeless}. This multiplicity becomes a problem when selection among models is effectively arbitrary and the differences between them, though invisible to aggregate metrics, are consequential at the level of individual predictions \citep{damour2022underspecification}: two models indistinguishable by AUC may nonetheless assign the same patient materially different risks, and therefore different clinical actions, depending on which solution the learning pipeline happened to find \citep{riley2023multiverse}. This phenomenon is closely related to the classical statistical notion of degrees of freedom, which formalizes how model complexity governs sensitivity to perturbations in the training data \citep{efron2004covariance}. 

The healthcare literature has long recognized that prediction models are sensitive to factors related to pipeline design and data, including variable selection \citep{steyerberg2005divergent} and training sample composition \citep{riley2023stability, Riley2019Prognosis}. This instability has been shown to manifest as miscalibration in new data and inconsistency in threshold-based classification decisions \citep{yusufujiang2026prediction, riley2023stability}. However, this body of work has focused primarily on the underparameterized regime ($p<n$), where the estimation procedure can be treated as approximately deterministic given the data. This means that all instability is effectively attributed to variability in the training sample. This framing becomes increasingly untenable for modern flexible models, which rely on stochastic optimization procedures involving random weight initialization and mini-batch sampling. These procedures introduce a second, independent source of variability: even when the training data, architecture, and hyperparameters are held fixed, retraining a model can produce materially different predictions for the same individual. Indeed, prior research \citep{pmlr-v193-lopez-martinez22a} has demonstrated this directly in a clinical risk stratification setting, showing that repeated training runs of the same deep learning model on identical data yielded significantly different orderings of high-risk patients, with direct implications for the allocation of scarce clinical resources. Similarly, prior research \citep{yusufujiang2026prediction} showed that models with comparable discrimination can produce markedly different individual risk estimates depending on modeling approach, with differences large enough to alter clinical decisions at the individual level. The present work extends this line of inquiry by introducing diagnostic tools that formally quantify and separate these two sources of individual-level prediction instability, evaluates the enitre risk spectrium, and compares across models of similar capacity.

In the overparameterized regime, the optimal solution is not a single coordinate but one of many equally performing points scattered across a complex, non-convex loss landscape \citep{Goodfellow2015Qualitative}. Which of these points a model converges to depends on the stochastic path taken during optimization. This instability has additional consequences beyond what will be discussed here, such as post-hoc explanations of individual predictions (e.g., feature importances) which in themselves can be unstable, meaning that attempts to interpret or audit the model may reflect an arbitrary draw from the solution space rather than any stable underlying logic \citep{riley2023stability}.By ignoring this additional source of stochasticity, current validation paradigms do not fully characterize the uncertainty inherent in the learning pipeline.

\section{Stability Metrics and Evaluation Framework}

We consider the standard supervised learning setting where observations $(X, Y)$ are drawn from an unknown joint distribution, yielding a dataset $D = \{(x_i, y_i)\}_{i=1}^n$. A standard training--evaluation paradigm partitions $D$ into a training set $D_{\text{train}}$ and a test set $D_{\text{test}}$. A fitted predictor $\hat{f}$ is obtained by applying a learning pipeline $\mathcal{A}$ to the training data. Modern learning pipelines often involve stochastic optimization, non-convex objectives, or both. To make this explicit, we write the fitted model from the $b-th$ run as
\[
\hat{f}^{(b)} 
= 
\mathcal{A}\big(D_{\text{train}}^{(b)}, f, \phi, \gamma),
\]
where $D_{\text{train}}^{(b)}$ denotes a (possibly perturbed) version of the training data, $f \in \mathcal{H}$ denotes a model specification within hypothesis space $\mathcal{H}$, $\phi$ is a fixed loss function, $\gamma$ denotes hyperparameter settings. Hyperparameter settings control both fixed (e.g., learning rate) and random (e.g., weight initialization) aspects of the learning pipeline.

Model architectures that are convex (e.g., linear models, logistic regression) tend to produce singular solutions if optimized using standard solvers (e.g., L-BFGS in scikit-learn) under a fixed dataset. Conversely, model architectures that are non-convex or overparameterized (e.g., overparameterized neural networks, random forest) can produce many plausible fitted solutions. Depending on the complexity of the data and the hypothesis space, different model architectures can produce similar out-of-sample performance. We formalize this collection of models as the set of competitive predictors. Let $\widehat{L}_{\text{test}}(f)$ denote the empirical test loss under $\phi$. For a tolerance $\varepsilon \geq 0$, we define the competitive set as:

\[
\widehat{\mathcal{R}}_{\text{set}}(\mathcal{H}, \varepsilon) 
=
\left\{
f \in \mathcal{H}
:
\widehat{L}_{\text{test}}(f)
\le
\min_{g \in \mathcal{H}} \widehat{L}_{\text{test}}(g)
+
\varepsilon
\right\}.
\]

This set represents a region of empirical indistinguishability in function space such that predictors within $\widehat{\mathcal{R}}_{\text{set}}$ are equivalent up to tolerance $\varepsilon$ with respect to aggregate test performance \cite{semenova2019simpler}. In our experiments, $\phi$ is the binary cross-entropy (BCE) loss, and $\mathcal{H}$ is defined by the model specifications described in Section~3.3.

\subsection{Individual-Level Evaluation Metrics}

While the competitiveness criterion ensures stability in the aggregate performance metrics, it does not guarentee consistency in the specific risk estimates $\hat{f}(x_i)$ assigned to individual observations. We analyze the resulting instability as arising from two distint pathways within the learning pipeline: (1) perturbations to the training data $D_{train}$ and (2) stochasticity inherent in the optimization routine. 

We represent the collection of predictions assigned to a specific individual across $B$ repeated instantiations of the learning pipeline $\mathcal{A}$ as:
\[
\widehat{P}_i := \{\hat{f}^{(b)}(x_i)\}_{b=1}^B,
\]
where each $\hat{f}^{(b)}$ satisfies the competitiveness criterion. Here, $\widehat{P}_i$ is a vector of $B$ scalar predictions corresponding to individual $x_i$. Collectively, the predictions for all $n$ test individuals across $B$ competitive runs form an $n \times B$ prediction matrix, $\mathbf{\hat{P}}$. If the dispersion for each row in $\mathbf{\hat{P}}$ is large, then the pipeline is individually unstable for that observation, even if the models are globally equivalent. We quantify this dispersion using both continuous and categorical stability metrics introduced below.

\subsubsection{Empirical Prediction Interval Width (ePIW)}

For each individual $x_i$, let $\widehat{P}_i$ denote the $B \times 1$ vector of predicted risks obtained across repeated instantiations of the learning pipeline. We seek to quantify the effective support of the central distribution of this prediction set, i.e., the range of risk values that may plausibly be assigned to a given individual under learning pipeline variability. 

For a nominal coverage level $1 - \alpha$, let $Q_{\alpha/2}(\widehat{P}_i)$ and $Q_{1-\alpha/2}(\widehat{P}_i)$ denote the lower and upper empirical quantiles of prediction set $\widehat{P}_i$. We define the empirical prediction interval width (ePIW) for individual $x_i$ as
\[
\mathrm{ePIW}(x_i, \alpha) 
\;=\; 
Q_{1-\alpha/2}(\widehat{P}_i) 
\;-\; 
Q_{\alpha/2}(\widehat{P}_i).
\]

In our experiments, we use $\alpha = 0.05$, corresponding to the central 95\% of the empirical prediction distribution. Larger values of $\mathrm{ePIW}$ indicate greater dispersion in the prediction space for a given individual. Conversely, $\mathrm{ePIW} = 0$ implies that the predicted risk is invariant across all applied realizations of the learning pipeline and is therefore deterministic relative to the chosen model class and training procedure. Conceptually, $\mathrm{ePIW}$ answers the question: \emph{What is the range of plausible risk scores an individual might receive from models with comparable aggregate performance?}

\subsubsection{Empirical Decision Flip Rate (eDFR)}

While ePIW captures dispersion in continuous risk estimates, many practical applications of machine learning rely on thresholded binary decisions. To assess the stability of such decisions, let $\tau \in (0,1)$ denote a fixed decision threshold. For each learning pipeline instance $b$, we define the induced binary decision for individual $x_i$ as:
\[
\hat{d}_i^{(b)}(\tau) = \mathbb{I}\{\hat{f}^{(b)}(x_i) \geq \tau\},
\]
where $\mathbb{I}\{\cdot\}$ denotes the indicator function.

We define the empirical decision flip rate (eDFR) for individual $x_i$ as the proportion of distinct pairs of pipeline instantiations that disagree in their induced binary decision:
\[
\mathrm{eDFR}(x_i, \tau) = \frac{2}{B(B-1)} \sum_{1 \leq b < b' \leq B} \mathbb{I} \left\{ \hat{d}_i^{(b)}(\tau) \neq \hat{d}_i^{(b')}(\tau) \right\}.
\]

The eDFR takes values in $[0,1]$, where values near 0 indicate highly stable binary decisions. Conceptually, $\mathrm{eDFR}$ answers the question: \emph{How often does an individual’s classification change when the model is refit?} This metric is particularly salient for clinical decision-making, as excessive classification flipping across competitive models undermines the perceived reliability and trustworthiness of algorithmic recommendations.

\subsection{Model Classes Considered}

We consider candidate predictors drawn from two broad model families: logistic regression and feedforward neural networks. These classes were selected to span a range of expressive capacity, from constrained parametric specifications to flexible, overparameterized architectures. The specific architectural choices are not the primary focus of this study; rather, these models serve as representative instances within a competitive set of predictors that achieve comparable aggregate out-of-sample performance.

All models considered produce predicted risks in the unit interval and are trained using binary cross-entropy (BCE) loss with $L_2$ regularization to prevent overfitting and ensure numerical stability. Model configurations are selected to satisfy the competitiveness criterion defined above. Table~\ref{tab:model_specs} summarizes the model specifications and fitting routines used in our simulation and clinical risk prediction experiments.

\begin{table*}[htbp]
\centering
\caption{Model specifications for candidate architectures used in simulation and GUSTO-I experiments. The parameter count $p$ reflects the total number of trainable weights and biases.}
\label{tab:model_specs}
\setlength{\tabcolsep}{10pt}
\renewcommand{\arraystretch}{1.2}
\begin{tabular}{lll r}
\toprule
Abbreviation & Model Family & Specification \& Fitting Routine & $p$ \\
\midrule
Log-LBFGS & Logistic Regression & Quasi-Newton (L-BFGS), $L_2$ reg. & 6 \\
Log-G     & Logistic Regression & Quasi-Newton (L-BFGS), $L_2$ reg. & 9 \\
Log-SGD   & Logistic Regression & SGD, $L_2$ reg. & 6 \\
Log-Poly  & Logistic Regression & L-BFGS, $L_2$ reg., polynomial features & 21 \\
NN-1L     & Neural Network & 1 hidden layer (width 40), ReLU, SGD & 281 \\
NN-2L     & Neural Network & 2 hidden layers (width 180), ReLU, SGD & 33,841 \\
NN-G      & Neural Network & 2 hidden layers (width 180), ReLU, SGD & 34,381 \\
\bottomrule
\end{tabular}
\end{table*}

\section{Experiments and Results}
\subsection{Experimental Design}

We evaluate individual-level prediction stability under two distinct sources of learning-pipeline variability: (i) variation in the training data and (ii) stochasticity in the optimization procedure. To isolate data-driven instability, we employ a subsampling approach. For the simulated experiments, training sets are drawn directly from the known data-generating population. For the clinical experiments, training sets of size $n_{\text{train}}$ are sampled without replacement from the larger GUSTO-I dataset. 

Across all experiments, each model specification is retrained $B=100$ times and evaluated on a respective fixed test set of size $n_{\text{test}} = 10,000$. For the simulation, this test set was drawn from the population distribution, while for the clinical experiments, it is a held-out portion of the GUSTO-I dataset. This fixed set serves as a proxy for a stable deployment population, allowing us to measure how individual risk estimates fluctuate across refits. We consider two training sample sizes, $n_{\text{train}} \in \{500, 5000\}$, to examine how instability evolves with increased data availability.

For optimization-driven variability, the training data remains fixed while random seeds are varied to control weight initialization and the sequence of mini-batch updates in stochastic gradient descent. All models are trained using binary cross-entropy (BCE) loss with $L_2$ regularization. For each experimental condition, we obtain an $n_{\text{test}} \times B$ prediction matrix $\mathbf{\hat{P}}$, from which individual-level ePIW and eDFR are computed.

\subsection{Results on Simulated Data}

We first study individual-level prediction instability in a controlled synthetic setting where the data-generating process is known and moderately misspecified. Data are generated from a logistic regression model with five predictors, where only three features carry a signal and two represent noise. This design yields a balanced classification task with an estimated decision threshold near $\tau \approx 0.53$.

\paragraph{Aggregate performance and competitiveness.}
Table~\ref{tab:agg_perf_sim_all} shows how all model specifications considered are competitive under standard aggregate performance metrics. At $n_{train} = 500$, BCE ranges from 0.520 (Log-SGD) to 0.536 (NN-2L) under resampled training data, and accuracy ranges from 72.4\% (Log-poly) to 73.8\% (Log-SGD). At $n_{train} = 5000$, these differences narrow further, with BCE ranging from 0.517 to 0.520 and accuracy from 74.0\% to 74.1\% across all model classes.  The variability across retraining runs for a given model architecture is negligible. From the perspective of conventional evaluation criteria, these models are therefore practically indistinguishable.

\begin{table}[h!]
\centering
\caption{Aggregate out-of-sample performance on simulated data (mean $\pm$ 1 SD across $B = 100$ retrainings) evaluated on Binary Cross-entropy (BCE) and accuracy}
\label{tab:agg_perf_sim_all}
\footnotesize
\setlength{\tabcolsep}{3pt}
\begin{tabular}{lcccccccc}
\toprule
& \multicolumn{4}{c}{\textbf{Resampled training data}} 
& \multicolumn{4}{c}{\textbf{Fixed training data}} \\
\cmidrule(lr){2-5} \cmidrule(lr){6-9}
& \multicolumn{2}{c}{$n_{\text{train}} = 500$} 
& \multicolumn{2}{c}{$n_{\text{train}} = 5000$}
& \multicolumn{2}{c}{$n_{\text{train}} = 500$}
& \multicolumn{2}{c}{$n_{\text{train}} = 5000$} \\
\cmidrule(lr){2-3} \cmidrule(lr){4-5}
\cmidrule(lr){6-7} \cmidrule(lr){8-9}
\textbf{Model} 
& BCE & Acc. 
& BCE & Acc.
& BCE & Acc.
& BCE & Acc. \\
\midrule
Log-LBFGS 
& $0.523 \pm .004$ & $73.4\% \pm .4$
& $0.518 \pm .001$ & $74.1\% \pm .1$
& $0.517 \pm .000$ & $74.1\% \pm .0$
& $0.519 \pm .000$ & $74.0\% \pm .0$ \\

Log-SGD   
& $0.520 \pm .003$ & $73.8\% \pm .3$
& $0.517 \pm .001$ & $74.1\% \pm .1$
& $0.517 \pm .000$ & $74.0\% \pm .1$
& $0.517 \pm .000$ & $74.0\% \pm .1$ \\

Log-poly  
& $0.534 \pm .006$ & $72.4\% \pm .6$
& $0.520 \pm .001$ & $74.1\% \pm .1$
& $0.527 \pm .001$ & $73.2\% \pm .0$
& $0.520 \pm .001$ & $74.0\% \pm .0$ \\

NN-1L     
& $0.529 \pm .004$ & $73.6\% \pm .3$
& $0.519 \pm .001$ & $74.1\% \pm .1$
& $0.527 \pm .001$ & $73.7\% \pm .1$
& $0.519 \pm .001$ & $74.3\% \pm .1$ \\

NN-2L     
& $0.536 \pm .006$ & $73.0\% \pm .5$
& $0.519 \pm .001$ & $74.1\% \pm .1$
& $0.536 \pm .002$ & $73.3\% \pm .1$
& $0.519 \pm .000$ & $74.0\% \pm .1$ \\
\bottomrule
\end{tabular}
\end{table}

\begin{figure*}[h!]
    \centering
    \includegraphics[width=\textwidth]{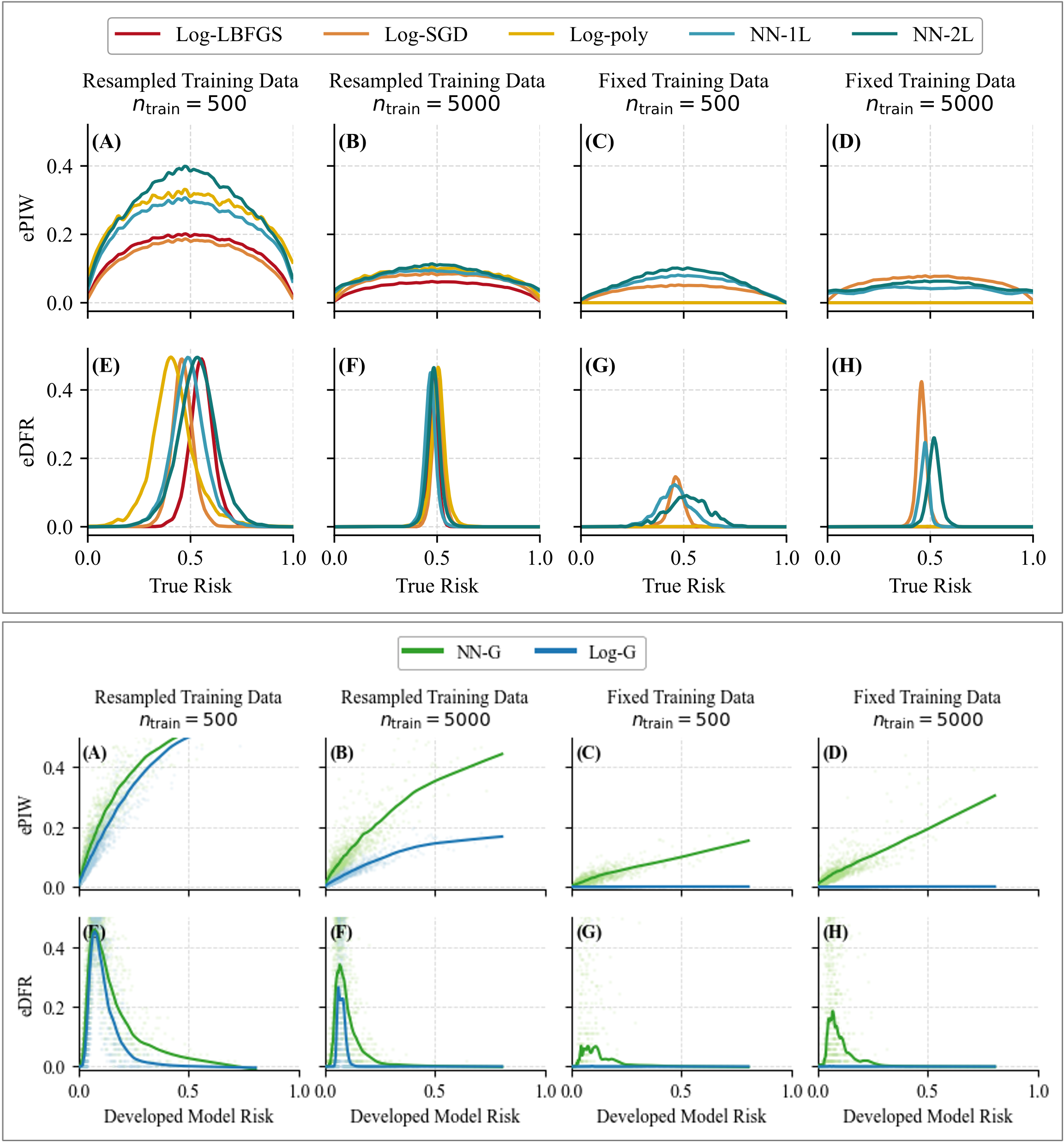}
    \caption{Individual-level prediction instability in simulation (top) and the GUSTO-I clinical dataset (bottom). Panels (A--D) report empirical prediction interval width (ePIW); panels (E--H) report empirical decision flip rate (eDFR). Columns correspond to resampled training data versus fixed training data (random-seed variation only), each evaluated at $n_{\text{train}} \in \{500, 5000\}$. In the simulated setting, instability is plotted as a function of true risk; in GUSTO-I, instability is plotted as a function of developed model risk. Across both settings, prediction dispersion is consistently larger for neural networks than for logistic regression and attenuates with increasing training sample size under resampled training data. Decision instability concentrates near clinical decision thresholds and diminishes with increased data and reduced sources of randomness.}
    \label{fig:sim_gusto_instability}
\end{figure*}

\paragraph{Individual-level prediction instability.}
Figure~\ref{fig:sim_gusto_instability} summarizes individual-level instability across the test population using ePIW and eDFR, and Tables~\ref{tab:epiw_all} and~\ref{tab:edfr_all} in the Appendix provide binned quantitative summaries. Model capacity and optimization procedures are substantially influential in shaping instability in the prediction space. In this data-generating condition, instability is concentrated among individuals with intermediate true risk, while predictions for low- and high-risk individuals are relatively more stable. As we discuss in the GUSTO-I analysis below, this concentration near the decision boundary is specific to the balanced class distribution of this simulation and where the decision threshold is defined relative to the risk estimates. In settings with greater class imbalance, the location of peak instability can shift accordingly.

At $n_{train} = 500$ under resampled training data, the magnitude of prediction differs substantially across model capacities and across the risk spectrum. In the [0.4, 0.5) true-risk bin (the region close to the decision threshold), Log-LBFGS produces an ePIW of 0.260, meaning that for a patient with true risk near
0.45, the predicted risk spans a 95\% empirical prediction interval of approximately 26 percentage points across retrainings.  The more flexible models exhibit considerably wider disperstions: ePIW reaches 0.389 for NN-1l and 0.491 for NN-2L. In practical terms, this two-layer neural network can assign the same patient a predicted risk anywhere in an interval nearly 49 percentage points wide, almost twice the spread of the more constrained logistical model. Increasing the training sample size to $n_{train} = 5000$ reduces instability under resampled training data for all models, but does not eliminate it. For individuals with true risk in [0.4, 0.5) peak ePIW falls to 8 percentage points for Log-LBFGS and 13--15 percentage points for NN-1L and NN-2L, with the relative ordering across model classes preserved.

As expected, for the fixed training data condition, models trained with a deterministic optimizer exhibit no sensitivity to the random seed. Stochastic optimization continue to exhibit meaningful instability, whereas deterministic optimization procedures yield stable individual predictions. Notably, for the highly flexible neural network models (NN-1L, NN-2L), instability induced by random seed variation alone is comparable in magnitude to that induced by resampling the training data. At $n_{train} = 500$, for individuals with true risk in [0.4, 0.5),  NN-1L and NN-2L produce risk scores that vary 0.107 and 0.137 percentage points wide. Unsurprisingly, the eDFR curves show that decision instability is sharply concentrated near the decision threshold. Outside of bins near the decision threshold, instability falls sharply toward zero for all models. 

\begin{figure*}[h!]
    \centering
    \includegraphics[width=\textwidth]{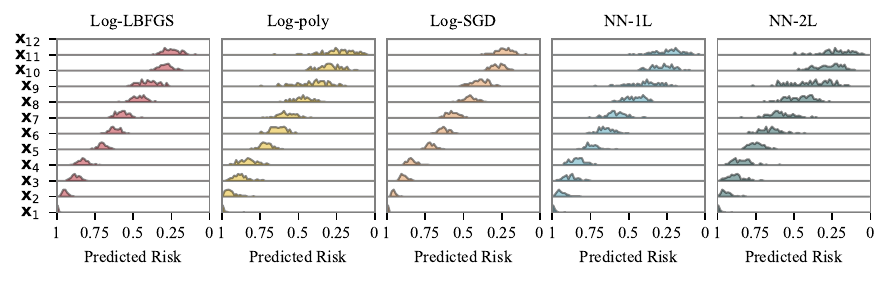}
    \caption {Individual-level prediction distributions across model families. Each panel displays the empirical distribution of predicted risks ($\widehat{P}_i$) for 12 representative individuals ($x_1$ through $x_{12}$) across $B$ repeated instantiations of the learning pipeline from resampling the training data. From left to right, the panels represent: (1) Log-LBFGS, (2) Log-poly, (3) Log-SGD, (4) NN-1L, and (5) NN-2L. While all models satisfy the competitiveness criterion, the ridge plots reveal varying degrees of individual-level dispersion, particularly as model complexity evolves to neural network architectures.}
    \label{fig:prediction_space_5panel}
\end{figure*}

\begin{figure}[h]
    \centering
    \includegraphics[width=0.95\textwidth]{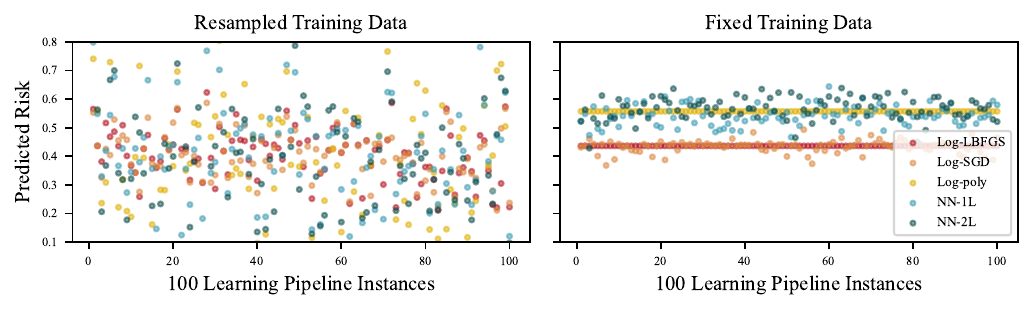}
    \vspace{-10pt}
    \caption{Prediction instability for a fixed out-of-sample individual (true risk $=0.381$) under $B=100$ learning pipeline (LP) instances with $n_{\text{train}}=500$ from simulation data. Each point denotes the predicted risk from one fitted model instance. \textbf{Left:} variability induced by resampling the training data. \textbf{Right:} variability induced by random seed initialization with fixed training data.}
    \label{fig:ind_instability_example}
\end{figure}

The prediction space can also be evaluated at the individual level to illustrate the heterogeneity of these distributions. Figure~\ref{fig:prediction_space_5panel} displays the empirical risk distributions for 12 representative individuals across $B = 100$ pipeline instantiations. Individual-level prediction behavior is further illustrated in Figure~\ref{fig:ind_instability_example}, which tracks predictions for a single out-of-sample individual (true risk $=0.381$) across 100 pipeline instantiations. Population-level summaries of prediction interval width, bias, and mean squared error are reported in Appendix~\ref{app:instability_summaries}.

\newpage 

\newpage 

\subsection{Results on Clinical Risk Prediction (GUSTO-I)}

We next assess individual-level prediction stability in a real clinical risk prediction task using the GUSTO-I trial on 30-day mortality following acute myocardial infarction \citep{gusto1993thrombolytic}. The outcome prevalence is 7\%, and the distribution of developed risk is highly skewed toward low values, with a clinically relevant decision threshold near $\tau \approx 0.07$. As shown in Table~\ref{tab:agg_perf_gusto_all}, all model specifications remain competitive under standard aggregate performance metrics (i.e., binary cross entropy and accuracy) across both training sample sizes. Similar to above, Figure~\ref{fig:sim_gusto_instability} shows substantial disparities in individual-level instability across model specifications. Consistent with our simulation findings, neural network architectures exhibit markedly greater dispersion in predicted risk ($\widehat{P}_i$) compared to logistic regression models under all experimental conditions.

\begin{table}[h!]
\centering
\caption{Aggregate out-of-sample performance on GUSTO-I (mean $\pm$ 1 SD across $B=100$ retrainings).}
\label{tab:agg_perf_gusto_all}
\footnotesize
\setlength{\tabcolsep}{4pt}
\begin{tabular}{lcccccccc}
\toprule
& \multicolumn{4}{c}{\textbf{Resampled training data}} 
& \multicolumn{4}{c}{\textbf{Fixed training data}} \\
\cmidrule(lr){2-5} \cmidrule(lr){6-9}
& \multicolumn{2}{c}{$n=500$}
& \multicolumn{2}{c}{$n=5000$}
& \multicolumn{2}{c}{$n=500$}
& \multicolumn{2}{c}{$n=5000$} \\
\cmidrule(lr){2-3} \cmidrule(lr){4-5}
\cmidrule(lr){6-7} \cmidrule(lr){8-9}
\textbf{Model}
& BCE & Acc.
& BCE & Acc.
& BCE & Acc.
& BCE & Acc. \\
\midrule

Log-G
& 0.22 $\pm$ .01 & 0.93 $\pm$ .00
& 0.21           & 0.93 $\pm$ .00
& 0.22           & 0.93 $\pm$ .00
& 0.21           & 0.93 $\pm$ .00 \\

NN-G
& 0.22 $\pm$ .01 & 0.93 $\pm$ .01
& 0.22           & 0.93 $\pm$ .00
& 0.23           & 0.92 $\pm$ .01
& 0.21           & 0.93 $\pm$ .01 \\

\bottomrule
\end{tabular}
\end{table}

\newpage

A critical distinction in the clinical setting is the localized nature of instability along the risk spectrum. While the simulation showed instability concentrated near the $\tau \approx 0.53$ boundary, the GUSTO-I analysis demonstrates that significant prediction dispersion (ePIW) occurs in the upper tail of the distribution. Because the clinical threshold ($\tau \approx 0.07$) is relatively low, much of this neural-network-induced variability occurs in regions that do not necessarily trigger a change in binary treatment assignments. For practitioners, this highlights a vital nuance, where elevated prediction instability does not strictly translate to decision instability (eDFR) if the dispersion is distal to the actionable threshold. Nevertheless, for high-risk patients, these fluctuations in mortality estimates can significantly undermine a clinician's confidence in the model's precision, even if the "high-risk" classification itself remains technically stable.

\section{Discussion}
Individual-level prediction instability is an often overlooked barrier to the deployment of machine learning in healthcare. It may not be immediately obvious why instability in individual-level predictions is problematic in practice. First, the variance of a risk estimate is often overlooked. Second, the only working optimization routine may also be contributing significantly to the instability. One might argue that variability across model fits is an inherent feature of predictive modeling, and that different models will naturally sort individuals differently. However, not only may the dependency of the predictions on the learning algorithm be a useful quality to know and communicate, but the sensitivity of individual predictions represents an additional and largely unexamined source of uncertainty that merits explicit consideration alongside standard measures of generalization performance. Statistical learning literature has long recognized that flexible models can be more sensitive to arbitrary changes in the training data \citep{efron2004covariance}, our findings highlight a distinct, complementary concern. This concern being that flexible models can also be more sensitive to arbitrary changes to the learning pipeline. We do not claim that optimization-induced instability is equivalent to overfitting in the traditional sense, rather, it represents a form of procedural arbitrariness where a patient’s treatment eligibility is determined as much by the random initialization of the model as by their clinical data. 

Our results reveal a tradeoff that mirrors the classical bias-variance decomposition but extends it into the domain of procedural consistency. More constrained models, such as logistic regression, exhibit low instability but may carry greater systematic bias relative to the true risk function, particularly in settings where the data-generating process is nonlinear. More flexible models can achieve lower bias at the cost of substantially greater instability. In settings where simpler models achieve comparable aggregate performance, as in our experiments, the instability cost of additional flexibility may not be justifiable. When there is a meaningful performance gap favoring the more flexible model, practitioners must weigh the gains in accuracy against the losses in individual-level consistency, taking into account the clinical consequences of inconsistent recommendations at the specific decision threshold in use.

The evaluation framework we propose provides a practical means of characterizing model sensitivity to the learning pipeline across the full range of model complexity. Stability diagnostics should be incorporated into routine model validation workflows rather than treated as optional post-hoc analyses. To support this, we propose the following checklist for evaluating whether a model's procedural consistency meets the standards required for high-stakes clinical deployment:

\begin{itemize}
    \item What is the dispersion of an individual's predicted risk across repeated retrainings, and does it vary across the risk spectrum?
    \item How frequently do perturbations in the learning pipeline cause a patient's clinical classification to flip?
    \item Are there specific subpopulations or risk strata where prediction instability is disproportionately high, and do these coincide with clinically actionable thresholds?
    \item Which specific modeling choices (e.g., architecture, optimizer, initialization) systematically induce greater instability, and can more stable alternatives achieve comparable predictive performance?
\end{itemize}

Finally, these findings help contextualize some skepticism among clinical stakeholders toward AI-assisted decision making \citep{Markowetz2024AllModelsWrong}. Such skepticism is often attributed to a lack of technical literacy \citep{Kumar2025Adoption} or resistance to change \citep{arvai2025concerns}. Our results suggest a more principled justification: individual-level predictions may be functionally inconsistent across retrainings depending on the model capacity. It is worth noting that the preferred balance between stability and flexibility may vary by clinical task. In settings where consistent, reliable predictions are highly valued (e.g., routine screening) more constrained models that sacrifice flexibility for procedural consistency may be preferable. In contrast, tasks that require discriminating among hard cases, where a more flexible model's ability to capture complex structure may outweigh its instability, may warrant a different tradeoff. Our framework does not prescribe a single answer, but rather makes the stability dimension of this choice explicit so that practitioners can reason about it deliberately rather than inadvertently.

\subsection{Limitations}
This work has several limitations. First, our proposed diagnostics require repeated model retraining, which may be computationally demanding for large-scale or resource-intensive models. Second, our analysis focuses on variability induced by resampling the training data and stochastic optimization procedures, and does not account for other sources of uncertainty such as label noise, distribution shift, or unmeasured confounding. Third, while we consider representative model classes spanning a range of expressive capacity, our findings may not generalize to all architectures or training regimes. Fourth, our stability metrics are empirical and descriptive; they do not provide formal uncertainty guarantees for individual predictions. Finally, our clinical evaluation is limited to a single risk prediction task and dataset, and broader validation across clinical domains and decision contexts is needed to assess the generality of these findings.

\section{Conclusion}

This work demonstrates that clinical prediction models with indistinguishable aggregate performance can nonetheless exhibit substantial, hidden variability at the level of individual predictions. This is variability that arises not only from the training data but from stochastic elements of the learning pipeline itself, such as random initialization and optimization noise. For high-capacity models, these optimization-induced fluctuations can be as impactful as resampling the entire training dataset, extending the classical understanding of model complexity and prediction stability into the overparameterized regime where the stochastic nature of optimization introduces an additional and largely unaudited source of uncertainty. To quantify this phenomenon, we introduced two model-agnostic diagnostics, empirical prediction interval width (ePIW) and empirical decision flip rate (eDFR). Taken together, these findings establish individual-level reliability as an essential and complementary criterion for clinical model evaluation. 

\bibliographystyle{unsrt}  
\bibliography{chil}  

\newpage 

\appendix

\section{Supplementary Results and Figures}
\label{app:instability_summaries}

\begin{table*}[h]
\centering
\caption{Empirical prediction interval width (ePIW) by true-risk bin across training conditions.}
\label{tab:epiw_all}
\small
\setlength{\tabcolsep}{6pt}
\begin{tabular}{lccccc}
\toprule
\textbf{True Risk (bin)} & \textbf{Log-LBFGS} & \textbf{Log-SGD} & \textbf{Log-poly} & \textbf{NN-1L} & \textbf{NN-2L} \\
\midrule
\multicolumn{6}{l}{\textbf{Resampled training data, $n_{\text{train}} = 500$}} \\
\midrule
$[0.0, 0.1)$ & 0.104 & 0.089 & 0.230 & 0.184 & 0.205 \\
$[0.1, 0.2)$ & 0.177 & 0.161 & 0.314 & 0.279 & 0.322 \\
$[0.2, 0.3)$ & 0.223 & 0.205 & 0.369 & 0.342 & 0.411 \\
$[0.3, 0.4)$ & 0.250 & 0.232 & 0.398 & 0.375 & 0.469 \\
$[0.4, 0.5)$ & 0.260 & 0.240 & 0.407 & 0.389 & 0.491 \\
$[0.5, 0.6)$ & 0.259 & 0.238 & 0.414 & 0.388 & 0.490 \\
$[0.6, 0.7)$ & 0.245 & 0.220 & 0.390 & 0.358 & 0.449 \\
$[0.7, 0.8)$ & 0.224 & 0.194 & 0.362 & 0.323 & 0.391 \\
$[0.8, 0.9)$ & 0.184 & 0.152 & 0.324 & 0.275 & 0.323 \\
$[0.9, 1.0]$ & 0.112 & 0.086 & 0.263 & 0.197 & 0.229 \\
\midrule
\multicolumn{6}{l}{\textbf{Resampled training data, $n_{\text{train}} = 5000$}} \\
\midrule
$[0.0, 0.1)$ & 0.033 & 0.048 & 0.073 & 0.073 & 0.076 \\
$[0.1, 0.2)$ & 0.056 & 0.082 & 0.104 & 0.096 & 0.101 \\
$[0.2, 0.3)$ & 0.070 & 0.102 & 0.120 & 0.121 & 0.126 \\
$[0.3, 0.4)$ & 0.078 & 0.111 & 0.129 & 0.130 & 0.145 \\
$[0.4, 0.5)$ & 0.081 & 0.114 & 0.130 & 0.128 & 0.154 \\
$[0.5, 0.6)$ & 0.081 & 0.116 & 0.131 & 0.124 & 0.154 \\
$[0.6, 0.7)$ & 0.077 & 0.111 & 0.125 & 0.116 & 0.138 \\
$[0.7, 0.8)$ & 0.070 & 0.101 & 0.116 & 0.105 & 0.118 \\
$[0.8, 0.9)$ & 0.058 & 0.082 & 0.099 & 0.088 & 0.096 \\
$[0.9, 1.0]$ & 0.034 & 0.047 & 0.066 & 0.074 & 0.078 \\
\midrule
\multicolumn{6}{l}{\textbf{Fixed training data, $n_{\text{train}} = 500$}} \\
\midrule
$[0.0, 0.1)$ & 0.000 & 0.034 & 0.000 & 0.038 & 0.036 \\
$[0.1, 0.2)$ & 0.000 & 0.059 & 0.000 & 0.064 & 0.066 \\
$[0.2, 0.3)$ & 0.000 & 0.073 & 0.000 & 0.085 & 0.099 \\
$[0.3, 0.4)$ & 0.000 & 0.080 & 0.000 & 0.102 & 0.124 \\
$[0.4, 0.5)$ & 0.000 & 0.083 & 0.000 & 0.107 & 0.137 \\
$[0.5, 0.6)$ & 0.000 & 0.084 & 0.000 & 0.107 & 0.136 \\
$[0.6, 0.7)$ & 0.000 & 0.079 & 0.000 & 0.101 & 0.121 \\
$[0.7, 0.8)$ & 0.000 & 0.072 & 0.000 & 0.088 & 0.093 \\
$[0.8, 0.9)$ & 0.000 & 0.061 & 0.000 & 0.063 & 0.057 \\
$[0.9, 1.0]$ & 0.000 & 0.035 & 0.000 & 0.028 & 0.022 \\
\midrule
\multicolumn{6}{l}{\textbf{Fixed training data, $n_{\text{train}} = 5000$}} \\
\midrule
$[0.0, 0.1)$ & 0.000 & 0.040 & 0.000 & 0.046 & 0.049 \\
$[0.1, 0.2)$ & 0.000 & 0.070 & 0.000 & 0.052 & 0.059 \\
$[0.2, 0.3)$ & 0.000 & 0.090 & 0.000 & 0.066 & 0.077 \\
$[0.3, 0.4)$ & 0.000 & 0.101 & 0.000 & 0.073 & 0.087 \\
$[0.4, 0.5)$ & 0.000 & 0.105 & 0.000 & 0.069 & 0.093 \\
$[0.5, 0.6)$ & 0.000 & 0.107 & 0.000 & 0.069 & 0.095 \\
$[0.6, 0.7)$ & 0.000 & 0.099 & 0.000 & 0.072 & 0.088 \\
$[0.7, 0.8)$ & 0.000 & 0.089 & 0.000 & 0.068 & 0.073 \\
$[0.8, 0.9)$ & 0.000 & 0.072 & 0.000 & 0.055 & 0.057 \\
$[0.9, 1.0]$ & 0.000 & 0.042 & 0.000 & 0.046 & 0.051 \\
\bottomrule
\end{tabular}
\end{table*}


\begin{table*}[htbp]
\centering
\caption{Empirical decision flip rate (eDFR) by true-risk bin across training conditions.}
\label{tab:edfr_all}
\small
\setlength{\tabcolsep}{6pt}
\begin{tabular}{lccccc}
\toprule
\textbf{True Risk (bin)} & \textbf{Log-LBFGS} & \textbf{Log-SGD} & \textbf{Log-poly} & \textbf{NN-1L} & \textbf{NN-2L} \\
\midrule
\multicolumn{6}{l}{\textbf{Resampled training data, $n_{\text{train}} = 500$}} \\
\midrule
$[0.0, 0.1)$ & 0.000 & 0.000 & 0.001 & 0.000 & 0.000 \\
$[0.1, 0.2)$ & 0.000 & 0.000 & 0.007 & 0.001 & 0.002 \\
$[0.2, 0.3)$ & 0.000 & 0.000 & 0.027 & 0.010 & 0.022 \\
$[0.3, 0.4)$ & 0.026 & 0.015 & 0.115 & 0.082 & 0.127 \\
$[0.4, 0.5)$ & 0.304 & 0.275 & 0.390 & 0.369 & 0.398 \\
$[0.5, 0.6)$ & 0.313 & 0.285 & 0.381 & 0.361 & 0.404 \\
$[0.6, 0.7)$ & 0.031 & 0.016 & 0.112 & 0.076 & 0.121 \\
$[0.7, 0.8)$ & 0.001 & 0.000 & 0.032 & 0.011 & 0.023 \\
$[0.8, 0.9)$ & 0.000 & 0.000 & 0.009 & 0.001 & 0.003 \\
$[0.9, 1.0]$ & 0.000 & 0.000 & 0.003 & 0.000 & 0.000 \\
\midrule
\multicolumn{6}{l}{\textbf{Resampled training data, $n_{\text{train}} = 5000$}} \\
\midrule
$[0.0, 0.1)$ & 0.000 & 0.000 & 0.000 & 0.000 & 0.000 \\
$[0.1, 0.2)$ & 0.000 & 0.000 & 0.000 & 0.000 & 0.000 \\
$[0.2, 0.3)$ & 0.000 & 0.000 & 0.000 & 0.000 & 0.000 \\
$[0.3, 0.4)$ & 0.000 & 0.000 & 0.005 & 0.001 & 0.001 \\
$[0.4, 0.5)$ & 0.104 & 0.130 & 0.168 & 0.140 & 0.155 \\
$[0.5, 0.6)$ & 0.116 & 0.142 & 0.183 & 0.149 & 0.173 \\
$[0.6, 0.7)$ & 0.000 & 0.000 & 0.003 & 0.000 & 0.001 \\
$[0.7, 0.8)$ & 0.000 & 0.000 & 0.000 & 0.000 & 0.000 \\
$[0.8, 0.9)$ & 0.000 & 0.000 & 0.000 & 0.000 & 0.000 \\
$[0.9, 1.0]$ & 0.000 & 0.000 & 0.000 & 0.000 & 0.000 \\
\midrule
\multicolumn{6}{l}{\textbf{Fixed training data, $n_{\text{train}} = 500$}} \\
\midrule
$[0.0, 0.1)$ & 0.000 & 0.000 & 0.000 & 0.000 & 0.000 \\
$[0.1, 0.2)$ & 0.000 & 0.000 & 0.000 & 0.000 & 0.000 \\
$[0.2, 0.3)$ & 0.000 & 0.000 & 0.000 & 0.002 & 0.004 \\
$[0.3, 0.4)$ & 0.000 & 0.000 & 0.000 & 0.017 & 0.022 \\
$[0.4, 0.5)$ & 0.000 & 0.059 & 0.000 & 0.076 & 0.062 \\
$[0.5, 0.6)$ & 0.000 & 0.062 & 0.000 & 0.099 & 0.083 \\
$[0.6, 0.7)$ & 0.000 & 0.000 & 0.000 & 0.022 & 0.038 \\
$[0.7, 0.8)$ & 0.000 & 0.000 & 0.000 & 0.002 & 0.007 \\
$[0.8, 0.9)$ & 0.000 & 0.000 & 0.000 & 0.000 & 0.000 \\
$[0.9, 1.0]$ & 0.000 & 0.000 & 0.000 & 0.000 & 0.000 \\
\midrule
\multicolumn{6}{l}{\textbf{Fixed training data, $n_{\text{train}} = 5000$}} \\
\midrule
$[0.0, 0.1)$ & 0.000 & 0.000 & 0.000 & 0.000 & 0.000 \\
$[0.1, 0.2)$ & 0.000 & 0.000 & 0.000 & 0.000 & 0.000 \\
$[0.2, 0.3)$ & 0.000 & 0.000 & 0.000 & 0.000 & 0.000 \\
$[0.3, 0.4)$ & 0.000 & 0.000 & 0.000 & 0.000 & 0.000 \\
$[0.4, 0.5)$ & 0.000 & 0.101 & 0.000 & 0.050 & 0.069 \\
$[0.5, 0.6)$ & 0.000 & 0.134 & 0.000 & 0.079 & 0.111 \\
$[0.6, 0.7)$ & 0.000 & 0.000 & 0.000 & 0.000 & 0.001 \\
$[0.7, 0.8)$ & 0.000 & 0.000 & 0.000 & 0.000 & 0.000 \\
$[0.8, 0.9)$ & 0.000 & 0.000 & 0.000 & 0.000 & 0.000 \\
$[0.9, 1.0]$ & 0.000 & 0.000 & 0.000 & 0.000 & 0.000 \\
\bottomrule
\end{tabular}
\end{table*}

\label{app:supp_figs}

\begin{figure*}[htbp]
    \centering
    \includegraphics[width=0.97\textwidth]{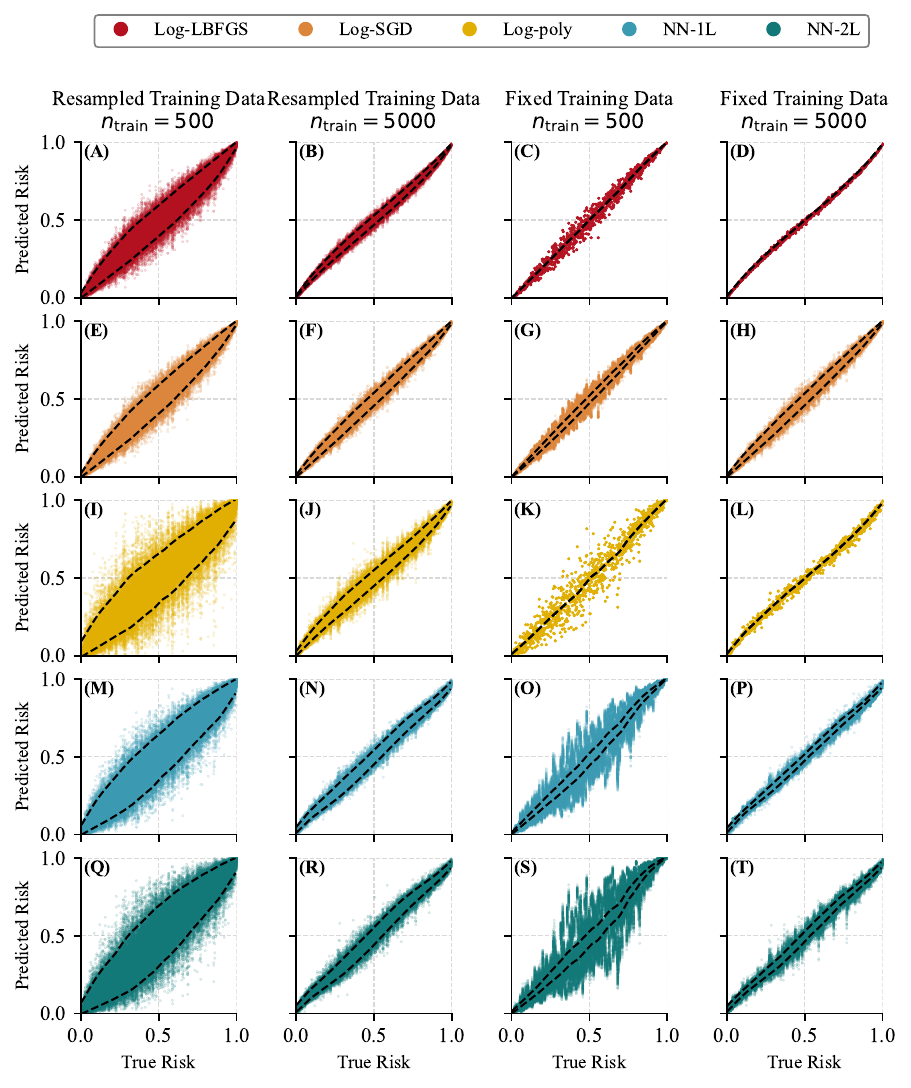}
    \caption{Predicted risk versus true risk across model classes, training sample sizes, and sources of stochasticity. Each row corresponds to a model specification (Log-LBFGS, Log-SGD, Log-poly, NN-1L, NN-2L). Columns report results under resampled training data and fixed training data (random-seed variation only), each evaluated at $n_{\text{train}} \in \{500, 5000\}$. Points represent individual-level predicted risks evaluated on a common test set across $B=100$ model retrainings. The dashed $45^\circ$ line indicates perfect calibration ($\widehat{p}=p_{\text{true}}$).}
    \label{fig:calibration_grid}
\end{figure*}

\begin{figure*}[t]
    \centering
    \includegraphics[width=\textwidth]{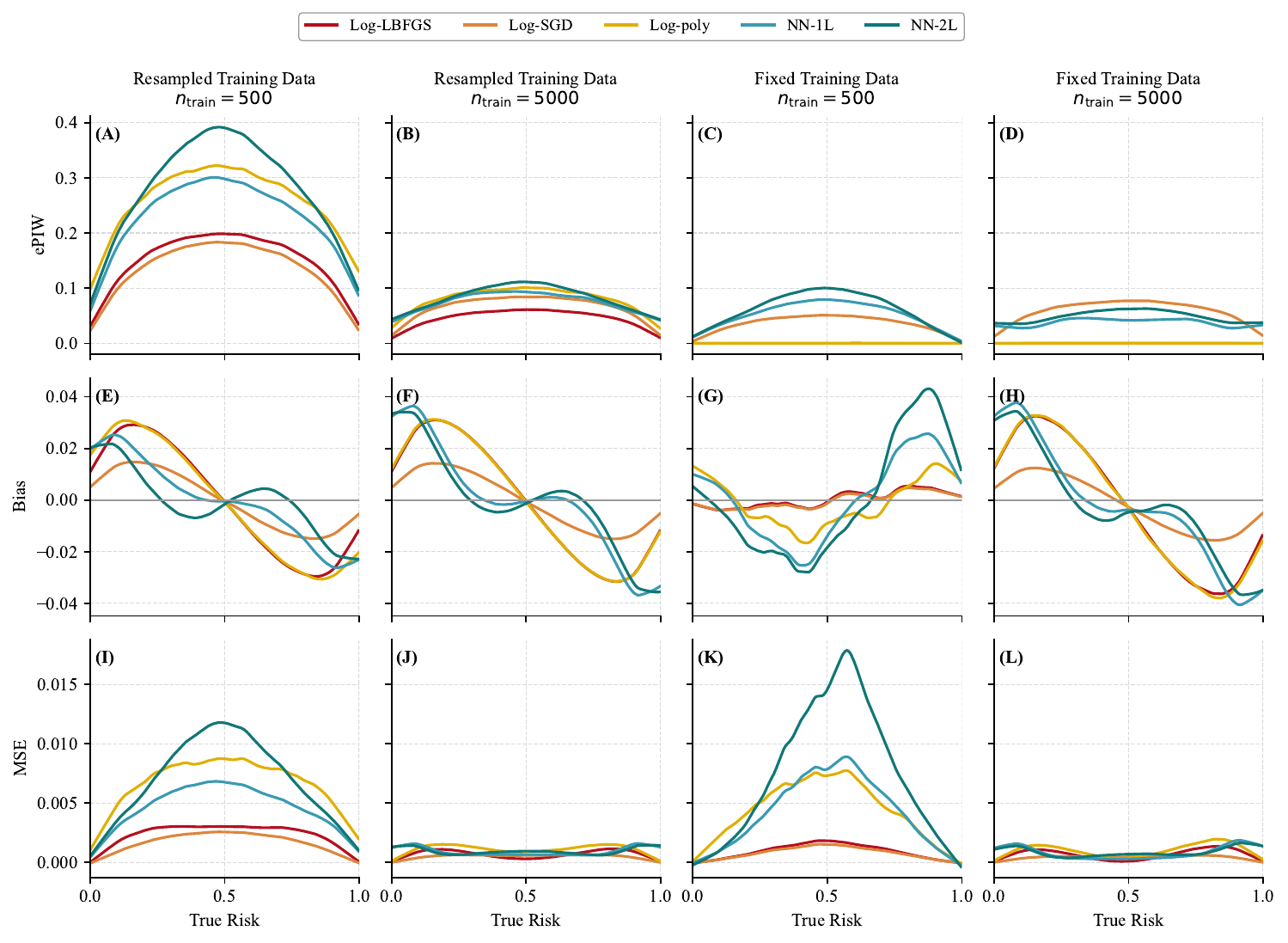}
    \caption{Individual-level prediction instability, bias, and mean squared error as functions of true risk in the simulated setting. Columns correspond to resampled training data and fixed training data, each evaluated at $n_{\text{train}} \in \{500, 5000\}$. Rows report empirical prediction interval width (ePIW; top), prediction bias (middle), and mean squared error (MSE; bottom), with LOESS smoothing applied to highlight systematic trends across the true-risk spectrum. Curves correspond to different model classes.}
    \label{fig:sim_epiw_bias_mse}
\end{figure*}

\end{document}